\documentclass{preprint}
\usepackage[backend=biber,style=vancouver]{biblatex}
\usepackage{array}
\newcolumntype{C}[1]{>{\centering\arraybackslash}m{#1}}
\usepackage{hyperref}
\usepackage{float}
\usepackage{amsmath}
\usepackage{rotating}

\begin{document}

\articletype{Article type}

\title{Online decoding of rat self-paced locomotion speed from EEG using recurrent neural networks}

\author{
Alejandro de Miguel$^{1,2,*}$\orcid{0000-0002-5327-6436},
Nelson Totah$^{3,4,5}$\orcid{0000-0002-0998-9770},
and Uri Maoz$^{1,2,6,7,8}$\orcid{0000-0002-7899-1241}
}

\affil{$^1$Schmid College of Science and Technology, Chapman University, Orange, California, USA}

\affil{$^2$LUCID-AI, Chapman University, Orange, California, USA}

\affil{$^3$Helsinki Institute of Life Science, University of Helsinki, Finland}

\affil{$^4$Neuroscience Center, University of Helsinki, Finland}

\affil{$^5$Faculty of Pharmacy, University of Helsinki, Finland}

\affil{$^6$Crean College of Health and Behavioral Sciences, Chapman University, Orange, California, USA}

\affil{$^7$Department of Biology and Bioengineering, California Institute of Technology, Pasadena, California, USA}

\affil{$^8$Anderson School of Management, University of California Los Angeles, Los Angeles, California, USA}

\affil{$^*$Author to whom any correspondence should be addressed.}

\email{demiguelgmez@chapman.edu}


\vspace{3mm}
\affil{Supplementary material for this manuscript is available online at 
\href{https://docs.google.com/spreadsheets/d/12vc-Sul2AjtMn3D7VbMtZNOV54iBW0T1/edit?usp=sharing&ouid=104978994468990027253&rtpof=true&sd=true}{this link}.}

\begin{abstract}
\textit{Objective.} Accurate and reliable neural decoding of locomotion holds promise for advancing clinical applications such as rehabilitation and prosthetic control, as well as for understanding neural correlates of action. Recent studies have demonstrated successful decoding of locomotion kinematics across species in motorized treadmill settings. However, efforts to decode locomotion speed directly and continuously in more natural contexts---where pace is self-selected rather than externally imposed---are scarce, and those that exist generally achieve only modest accuracy and require intracranial implants. Here, we aim to decode self-paced locomotion speed non-invasively and continuously using cortex-wide EEG recordings from rats. \textit{Approach.} We introduce an asynchronous brain--computer interface (BCI) that processes a continuous stream of 32-electrode skull-surface EEG recordings (0.01--45~Hz) to decode instantaneous speed readouts from a non-motorized treadmill during self-paced locomotion in head-fixed rats. Using recurrent neural networks and a large dataset comprising over 133~h of recordings, we trained decoders to map ongoing EEG activity to treadmill speed. \textit{Main results.} Our decoding methodology achieves a correlation of 0.88 ($R^2 = 0.78$) for speed, primarily driven by visual cortex electrodes and low-frequency ($< 8$~Hz) oscillations. Moreover, pre-training on a single recording session permitted decoding on other sessions from the same rat, suggesting the presence of uniform neural signatures of locomotion that generalize across sessions but fail to transfer across animals. Finally, we found that cortical states not only carry precise information about the current speed, but also about future and past dynamics, extending up to 1000~ms. \textit{Significance.} These findings demonstrate that self-paced locomotion speed can be decoded accurately and continuously from non-invasive, cortex-wide EEG. Our approach may provide a useful framework for developing high-performing, non-invasive BCI systems for locomotion and contribute to understanding distributed neural representations of action dynamics.
\end{abstract}

\section{Introduction}
The ability to decode locomotion from neural signals with high temporal precision is central to understanding how the brain integrates motor, sensory, and cognitive processes during movement, thereby providing a foundation for designing BCIs that can accurately translate neural activity into motor control. Among the many aspects of locomotion, speed is a particularly informative variable. From a neuroscience perspective, it reflects moment-to-moment motor output, integrates internal goals and sensory feedback, and shapes downstream spinal and muscular dynamics. From an engineering perspective, its temporal derivative corresponds to acceleration and, in principle, to force given mass (Newton’s second law). Yet it remains unclear how precisely cortical activity alone can track and predict continuous locomotion speed over behaviourally rich timescales, especially in self-paced conditions.

Work across species has demonstrated that locomotor states can be read out from diverse neural signals, primarily using invasive recordings (\autoref{tab:locomotion_summary}). In rodents, locomotor variables such as gait phase, discrete locomotor tasks, treadmill speed and slope, and continuous walking speed have been decoded from primary motor cortex, cerebellum, epidural EEG, and widefield calcium imaging  \cite{rigosa2015,mirfathollahi2022,muzzu2018,li2021,benisty2023}. Reported performances reach $\sim 90\%$ accuracy for classifying gait phases and discrete locomotor tasks and for distinguishing pre-set treadmill speeds and slope angles. In the continuous case, correlation coefficients approach 0.80 for linear decoding of self-paced walking speed from small cerebellar ensembles composed of the best-performing neurons (as identified by post-hoc selection), whereas dorsal cortex-wide optical imaging explains a more modest fraction of speed variance (R² $\approx 0.45$). 

\begin{table}[htbp]
\centering
\small
\renewcommand{\arraystretch}{1.2}
\begin{tabular}{
    C{3.2cm}
    C{1.7cm}
    C{2.2cm}
    C{1.7cm}
    C{2.0cm}
    C{2.0cm}
}
\hline
\textbf{Study} &
\textbf{Model} &
\textbf{Neural Recordings} &
\textbf{Locomotion Control} &
\textbf{Target} &
\textbf{Decoding Method} \\
\hline

Rigosa et al. \cite{rigosa2015} &
Rodent (rat) &
Invasive, M1 (single- and multi-unit) &
Enforced and self-paced &
Gait phase, discrete locomotion tasks &
Classification \\
\hline

Mirfathollahi et al. \cite{mirfathollahi2022} &
Rodent (rat) &
Invasive, M1 (LFP) &
Enforced &
Speed and slope &
Classification \\
\hline

Muzzu et al. \cite{muzzu2018} &
Rodent (mouse) &
Invasive, cerebellum (single-unit) &
Self-paced &
Speed &
Regression \\
\hline

Li et al. \cite{li2021} &
Rodent (rat) &
Invasive, cortex-wide (implanted EEG/epidural) &
Self-paced &
Slope &
Classification \\
\hline

Benisty et al. \cite{benisty2023} &
Rodent (mouse) &
Invasive (cranial window), dorsal cortex (widefield calcium imaging) &
Self-paced &
Speed &
Regression \\
\hline

Fitzsimmons et al. \cite{fitzsimmons2009} &
Non-human primate (rhesus macaque) &
Invasive, M1 and S1 (single- and multi-unit) &
Enforced &
Speed &
Regression \\
\hline

Foster et al. \cite{foster2014} &
Non-human primate (rhesus macaque) &
Invasive, PMd (multi-unit) &
Enforced &
Speed &
Classification \\
\hline

Capogrosso et al. \cite{capogrosso2016} &
Non-human primate (rhesus macaque) &
Invasive, M1 (multi-unit) &
Enforced &
Gait phase &
Classification \\
\hline

Lisi and Morimoto \cite{lisi2015} &
Human &
Non-invasive, cortex-wide (EEG) &
Enforced &
Speed change &
Classification \\
\hline

Wang et al. \cite{wang2021} &
Human &
Invasive, M1 (ECoG) &
Enforced &
Walking state, step rate &
Classification and regression \\
\hline
\end{tabular}
\caption{\textbf{Summary of relevant literature directly addressing decoding of locomotion speed or related locomotor variables from neural signals, across species.} Each study listed includes speed decoding as a primary or secondary objective.}
\label{tab:locomotion_summary}
\end{table}

\newpage
In non-human primates, intracortical recordings from motor, somatosensory, and premotor areas support decoding of treadmill speed and gait speed categories, with correlations of 0.24–0.42 for linear reconstruction of enforced treadmill speed and $\sim 85\%$ accuracy for classifying slow versus fast walking, and gait phase detection accuracies up to 97\% in pre-lesion conditions when coupled to wireless brain–spinal interfaces \cite{fitzsimmons2009,foster2014,capogrosso2016}. In humans, EEG and ECoG studies have reported  $\sim 73\%$ cross-validated accuracy (AUC 0.80–0.89) for distinguishing steady-state walking from gait speed transitions, and near-perfect classification of walking states (99.8\% accuracy) together with step-rate regression correlations of $\sim 0.93$—although the latter relied on a causal Bayesian filter whose output lagged true cadence by roughly 5.5 s, requiring offline temporal alignment for performance evaluation \cite{lisi2015,wang2021}. Collectively, these findings indicate that latent neural representations of locomotor states are accessible across species and recording modalities, and they establish locomotion decoding as a promising avenue for both basic and translational research.

Despite this progress, several important limitations remain. Many studies, including additional work not focused explicitly on speed \cite{barroso2019,liang2023,xing2019}, prioritize limb kinematics, joint angles, or discrete gait phases as decoding targets—variables that often exhibit large variance, align naturally to trial-based designs, and have direct clinical relevance for restoring stepping \cite{capogrosso2016,lorach2023}. In contrast, continuous locomotion speed is frequently treated as a secondary outcome or is constrained by experimental designs that enforce a small set of treadmill speeds and inclines. Such designs reduce within-session variability in pace and favour classification of discrete gait states over regression of speed, making it difficult to assess how accurately neural activity tracks the full range of voluntary locomotor dynamics. Importantly, experimental designs that impose externally controlled speeds and rigid block structures restrict the subject’s autonomy over locomotion speed, casting doubt on the ability to decode performance in self-paced scenarios that involve additional planning and sensory feedback processes.

Similarly, research has centered on specific brain regions, with most studies predominantly examining signals from the motor cortex. This focus largely excludes potential contributions from other cortical areas involved in multisensory integration, which could provide complementary information and novel results into locomotion decoding.

Building on these observations, our study examines the precision with which self-paced locomotion speed can be continuously decoded from non-invasive, cortex-wide EEG in rats. Using a large dataset of over 133 hours of 32-channel EEG recordings from 14 rats that captured a wide range of self-controlled speeds on a non-motorized treadmill, we investigate how cortically-distributed signals support high-performing, session-wide decoding through non-linear, end-to-end deep-learning models. Rather than focusing on trial-based events or fixed, imposed speeds, our approach emphasizes continuous online performance under more natural conditions to measure cortical contributions and evaluate generalization across sessions and subjects in a behaviorally rich setting. Additionally, we assess the extent to which neural activity encodes not only the current, but also future and past treadmill speeds—a direction that, to the best of our knowledge, remains unexplored, and may offer insights into anticipatory neural dynamics underlying voluntary locomotion control.

\section{Methods}

\subsection{Subjects, surgery, and behavioral task}

The data used in this study was collected in a previous experiment involving 14 male rats each implanted with permanent electrodes to record EEG from the skull surface \cite{doutel2024}. Rats were pair-housed. Experiments were conducted during the active phase. Rats were housed in a reverse light cycle (07:00 lights off, 19:00 lights on). Experiments were performed in a walk in Faraday cage with DC-powered red ($>$ 690 nm peak wavelength) dim illumination.

All procedures were carried out after approval by local authorities and in compliance with the German Law for the Protection of Animals in experimental research (Tierschutzversuchstierverordnung) and the European Community Guidelines for the Care and Use of Laboratory Animals (EU Directive 2010/63/EU).

The animals were originally trained to perform a discrimination task based on brief visual stimuli by running past a distance threshold in response to one stimulus and sitting immobile in response to the other stimulus. Variable inter-trial intervals (2 to 3 sec), which were restarted if the rat ran in the 500 ms prior to stimulus onset, yielded epochs of task-unrelated locomotion. Since a distance threshold was used as the instrumental response in the task, the responses varied in velocity, duration, and latency to the visual stimulus. Collectively, the dataset captured a wide range of self-paced locomotion behaviors. More details on the behavioral task and surgery are available in \cite{doutel2024}.

\subsection{EEG and treadmill signal acquisition}

The EEG array (Neuronexus, USA) consisted of 32 electrodes permanently fixed to the surface of the skull. The array was aligned to bregma so that signals could be assigned to the immediately underlying brain region and compared across subjects. Each brain region was identified using a rat brain atlas \cite{paxinos2013}. A chlorided silver wire laying over the dural surface above the cerebellum was used as the ground electrode.

Wideband (0.1 Hz to 6 kHz) EEG signals were amplified using a pre-amplifying headstage (HS-36, Neuralynx, USA) and digitized (Digital Lynx SX, Neuralynx, USA). Signals were recorded at a sampling rate of 32 kHz. A lowpass filter (2nd order Butterworth) was applied offline at 45 Hz (MATLAB, filtfilt function). After filtering, the signals were downsampled offline to 100 Hz.

The angular position of the treadmill was recorded as an analog signal, alongside the continuous EEG signals, at a sampling rate of 32 kHz. Treadmill speed was calculated offline at a sampling rate of 100 Hz to create paired samples of EEG and speed.

\subsection{Inclusion criteria}

A session qualified for the study when all the following predefined conditions were met: (i) all 32 EEG channels were functional; and (ii) the treadmill speed trace exhibited sufficient variability to permit meaningful decoding. Locomotion variability was measured using the interquartile range (IQR) of treadmill speeds within each session—with absolute treadmill speed values ranging approximately from 0 to 7 (in arbitrary units). Sessions with an IQR lower than or equal to 0.46, corresponding to the 10th percentile of the IQR distribution, were considered to have low variability and were excluded from the analysis. There were initially 276 recording sessions across 14 rats (19.7 ± 12.5, mean ± STD, sessions per rat; range 2–37). After applying the inclusion criteria, 225 sessions (81.5\%) remained, still representing all 14 animals (16.1 ± 10.8 sessions per rat; range 2–34). As a result, the final dataset contains over 133 hours of continuous EEG and treadmill data, adding up to nearly 48 million paired neural samples and speed measurements.

\subsection{Neural decoding framework}

\begin{figure}[htbp]
 \centering
        \includegraphics[width=1\textwidth]{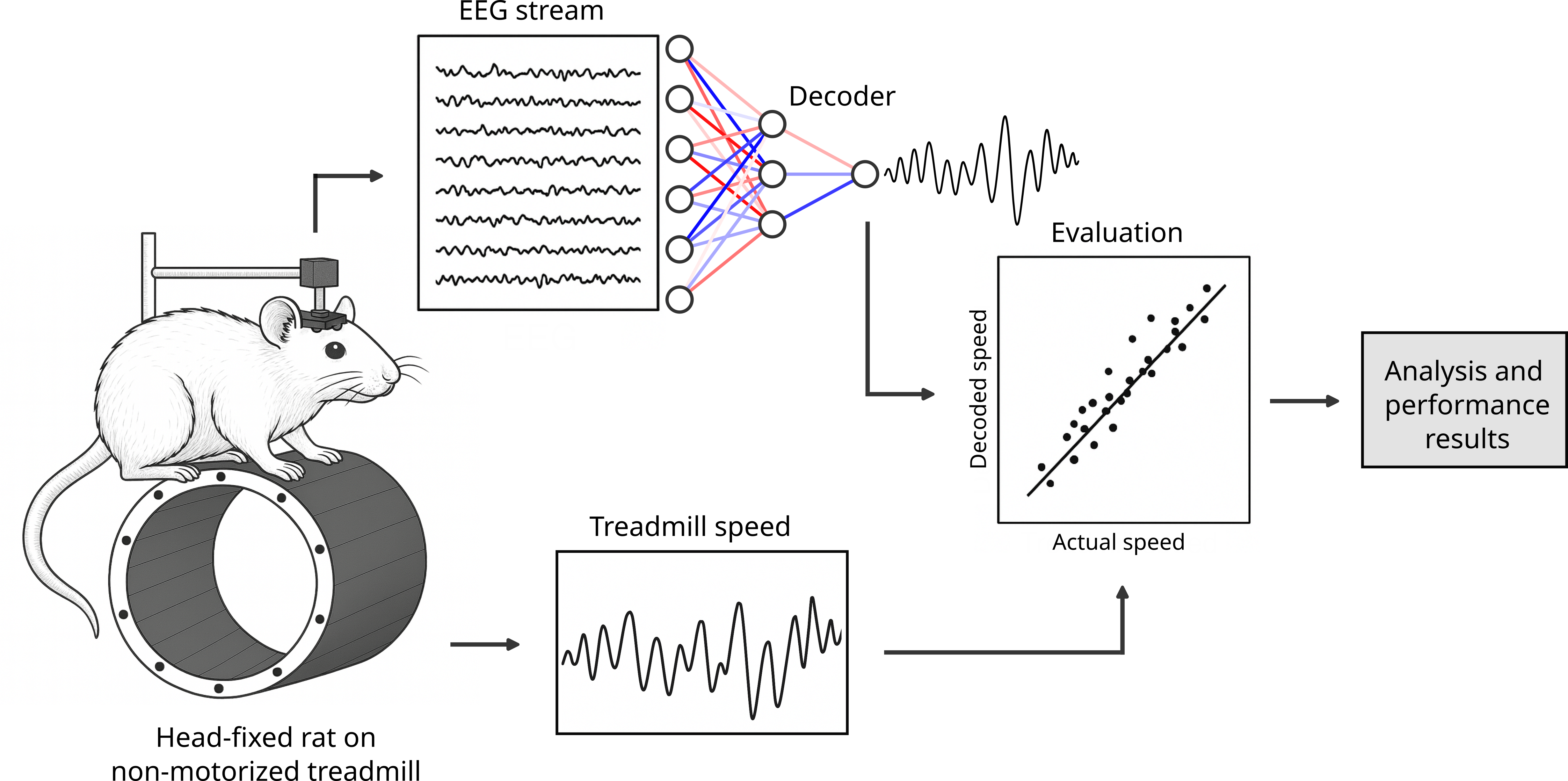}
 \caption{\textbf{Conceptual diagram of the decoding strategy.} This schematic is provided for visualization purposes only.}
\label{fig:diagram_rat}
\end{figure}

The primary goal of our decoding framework was to establish a direct, end-to-end mapping from neural activity to treadmill speed using supervised learning, formulated as a regression task. The decoders operate on a pseudo-real-time (offline simulation) continuous stream of z-scored neural data using a sliding window of 200 ms of EEG (32 channels × 20 time steps) to estimate instantaneous treadmill speed asynchronously. The window advances sample by sample, with a 10 ms stride, producing continuous outputs throughout a session. The window size was selected based on initial exploratory analyses, which showed that larger windows did not substantially improve decoding performance while increasing the dimensionality of the feature space and computational overhead. A schematic of the decoding pipeline is shown in \autoref{fig:diagram_rat}.

We trained and tested 5 different regression models: Linear Regression, Random Forest, Feed-Forward Neural Network, a Recurrent Neural Network (RNN) with Long Short-Term Memory (LSTM) units, and an Encoder-Only Transformer with self-attention

For the Linear Regression, Random Forest, and Feed-Forward Network, the 32×20 matrix from each window was flattened into a 640-dimensional vector, treating each channel-time-point pair as an independent feature.

The RNN input was structured as a 2-dimensional array with shape 20×32, where each row corresponds to a single time step of the window size, and each column represents an EEG channel. This time-major format allows the RNN to read the input row by row—one time step at a time—so that at each step, the model receives a 32-dimensional vector capturing the full spatial representation of neural activity at that moment. This enables the LSTM units to learn both temporal dynamics and cross-channel dependencies across the input window. At the end of the network, fully-connected, dense layers are applied to map the extracted features to a continuous speed estimate.

Similarly, the Encoder-Only Transformer processes sequences of 20 time steps where each step represents a token as a 32-dimensional EEG snapshot, thus the number of tokens equals the sequence size. Each token is projected to a larger number of dimensions analogous to a word-embedding layer in Natural Language Processing. We use a full bidirectional self-attention mechanism where every token attends to every other token as opposed to causal masking—in case the sequence of neural activity has meaning beyond the linearity of time. With the context-aware features, we apply a 1-dimensional convolution to look for relevant fluctuations that are meaningful in the context. Fully-connected, dense layers then produce the final output. 

Given the size of the dataset and the complexity of the deep-learning models evaluated, performing a full hyperparameter search using cross-validation is typically not feasible due to computational and time constraints. Instead, it is common practice to propose model configurations that have shown promise during an initial exploratory phase of the project. This approach has been widely adopted since the early days of deep-learning and remains standard in deep-learning-based EEG research \cite{roy2019}.

During training, we minimized the mean squared error (MSE) and used a hold-out validation set to ensure our models generalized well to unseen data. This allowed us to monitor validation loss, save the weights of the best-performing models, and implement early stopping when training loss started to diverge from validation loss—a common sign of overfitting and diminishing returns. In practice, a single validation set is often sufficient for large datasets, as its sample statistics closely approximate population-level properties, thereby reducing variance. Once again, in such cases, the benefit of cross-validation in reducing variance from random splits becomes less critical, while incurring a substantial increase in computational burden (see Section~\ref{model_selection} below for our cross-validation procedure). Further implementation details, including optimizer choices, regularization methods, and proposed model architectures, are provided in a publicly available \href{https://github.com/alejandrodmg/rat-locomotion-neural-decoding}{code repository}.

\subsection{Data analysis}

The relationship between EEG activity and locomotion was examined across a range of complimentary decoding experiments. For ease of evaluation across the different models, we computed the Pearson correlation coefficient (r) and coefficient of determination (R²) between the decoded speeds and the actual recordings. This provides a normalized measure of precision, allowing us to compare performance across animals, sessions and decoding configurations.

\subsubsection{Session-wide cross-validation}
\label{model_selection}
To measure the ability of the proposed models to decode treadmill speed, we trained them on continuous single-subject single-session neural activity, respecting the time-ordered nature of the data. For each session, the dataset was partitioned sequentially: the first 80\% was used for training, the next 10\% for validation, and the final 10\% for testing. The model that achieved the highest median correlation and R² on the held-out test set across all subjects and sessions was then selected for subsequent analysis steps. Within each session, EEG samples were z-scored channel-wise using the mean and standard deviation computed from the training set, and these same parameters were applied to the validation and test sets. This standardization approach was consistently used in all downstream analyses.

\subsubsection{Transfer learning procedure}
\label{transfer_learning}

\begin{figure}[htbp]
 \centering
        \includegraphics[width=1\textwidth]{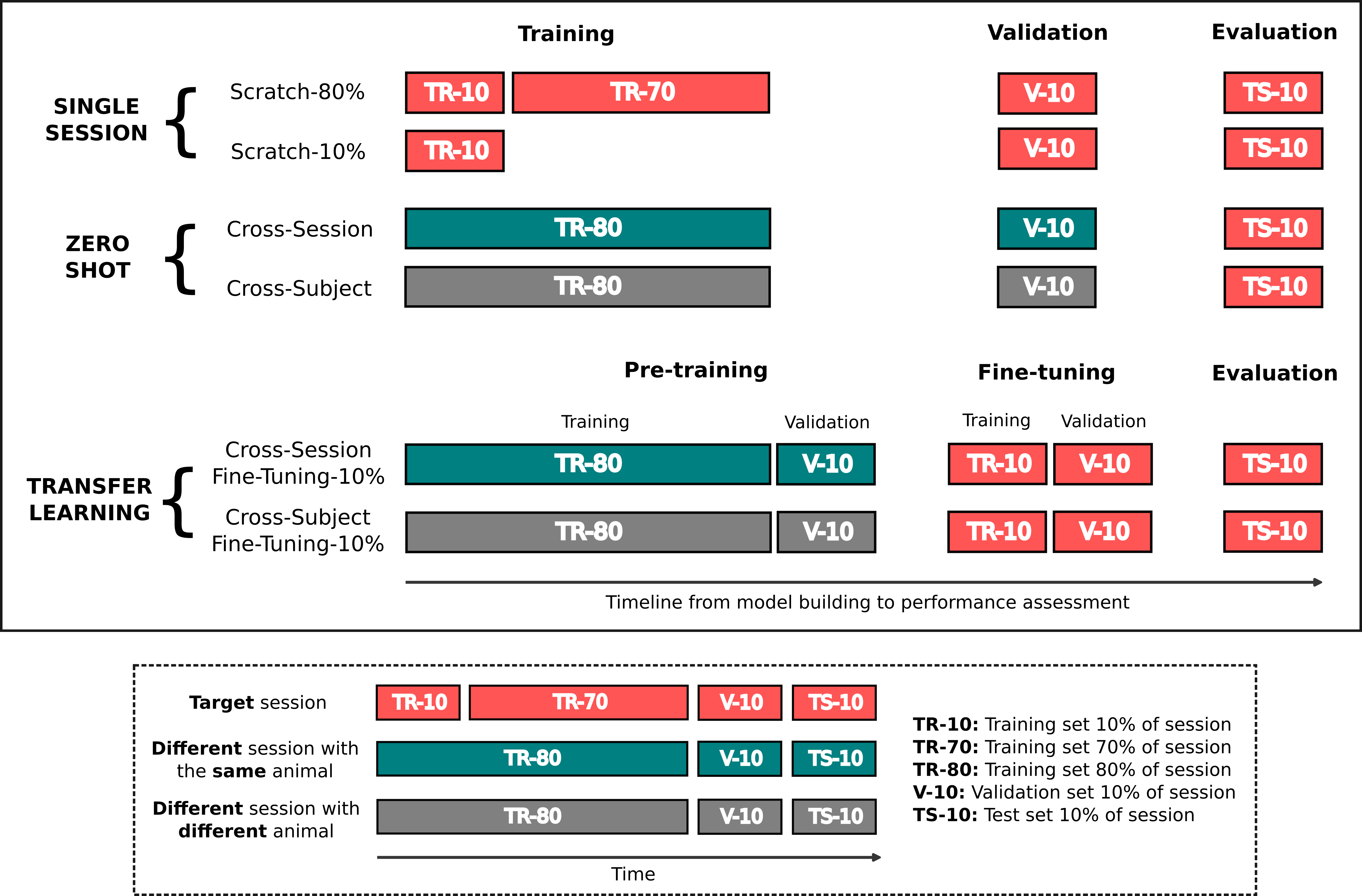}
 \caption{\textbf{Schematic of training strategies and data splits for assessing generalization of locomotion-related neural signatures across sessions and animals.}Models were evaluated under three strategies: single-session training, where decoders were trained from scratch on either 80\% or 10\% of the target session; zero-shot generalization, where a model trained on a different session with the same animal (cross-session) or a different subject (cross-subject) was directly applied to the target session without further training; and transfer learning, where pre-trained cross-session or cross-subject models were fine-tuned using only 10\% of the target session. Each bar represents a data partition: training (TR), validation (V), and test (TS), with numbers indicating the proportion of session data used.}
\label{fig:transfer_learning_diagram}
\end{figure}

We investigate the generalizability of locomotion-related neural patterns by directly applying pre-trained models on a single session (using 80\% of the data for training, as in Section~\ref{model_selection}) to all other sessions with the same subject (cross-session) and to sessions from different subjects (cross-subject). This zero-shot decoding attempt was compared to models trained from scratch on 80\% of the target session (best-case scenario baseline), only the first 10\% of a session—a common setup in real-time BCI applications—and to a transfer learning scenario, in which the cross-session and cross-subject pre-trained models were fine-tuned using the same 10\% of the new session. This fine-tuning approach differs from models trained from scratch in that the decoder is initialized using a set of weights learned during pre-training, which extracted features related to locomotion, and only the final fully connected layers of the model—responsible for mapping these features to treadmill speed—are retrained on the beginning of a new session to adapt to the current conditions. To ensure consistency with single-session evaluations, all models were tested on the last 10\% of each session, regardless of the type of learning. See \autoref{fig:transfer_learning_diagram} for a detailed schematic of the training strategies—single-session, zero-shot, and transfer learning—and their corresponding data splits.

The structure of this analysis determines the total number of train--test evaluations. 
For single-session models, each session contains both the training and test splits, 
producing exactly one performance estimate (correlation and $R^2$) per session; thus, 
the total number of evaluations equals the total number of sessions in the dataset 
(as in Section~\ref{model_selection}). For zero-shot cross-session decoding, let $K$ be the number of 
rats and $N_i$ the number of sessions for rat $i$. For each rat, every session is used 
once for training the decoder, which is then tested on all other sessions from the same 
rat, producing $\sum_{i=1}^{K} N_i (N_i - 1)$ evaluations or unique train--test combinations. For zero-shot cross-subject decoding, 
each decoder trained on a single session is instead tested on all sessions from 
different rats. If $N_{\text{total}}$ denotes the total number of sessions across all 
rats, this yields $\sum_{i=1}^{K} N_i (N_{\text{total}} - N_i)$ evaluations or unique train--test combinations.

In both cross-session and cross-subject settings (and their corresponding fine-tuning 
variants, which require calibrating the pre-trained models on 10\% of every new 
session), multiple models are evaluated on the same target session. To obtain a single, 
comparable performance estimate per session, we compute the median correlation and 
$R^2$ across all models evaluated on that target session. Consequently, each session 
contributes one correlation and one $R^2$ value for each training strategy, matching 
the single-session evaluation structure.

\subsubsection{Spatial and spectral analyses of the decoding signal}
\label{spatial_and_spectral}

The EEG electrodes were placed to be above specific brain regions (medial prefrontal, somatomotor, motor, and visual). Hence, to identify the source electrode of the decoded speed, we performed region-specific and pairwise comparisons of signals from different electrodes (spanning bilaterally from the most anterior frontal cortex regions to the most posterior visual cortical regions) with the aim of revealing the areas that carry the most salient information for decoding locomotion. In each case, models were trained, validated and tested on single sessions using only a subset of electrodes (e.g., on motor cortex only; or on combined visuo-motor electrodes, etc). Similarly, a frequency band analysis was conducted by evaluating model performance on isolated frequency ranges to determine relevant EEG bands driving decoding. Band-pass filtering was performed using zero-phase, 4th-order Butterworth filters to extract the following bands: delta, 1–4 Hz, theta, 4–8 Hz, alpha,  8–12 Hz, beta, 12–30 Hz, and gamma 30 Hz highpass. Note that in both the spatial and spectral analyses, each model was trained, validated, and evaluated using the same single-session data splits (80/10/10) described in Section~\ref{model_selection}, producing exactly one performance estimate (correlation and R²) per session.

To visualize the relationship between EEG spectral content and locomotion speed, we computed power spectral density (PSD) estimates across speed percentiles. For each session, the continuous speed trace was divided into deciles (0--10\%, \dots, 90--100\%) using within-session percentiles. For each decile, we extracted the EEG samples whose simultaneously measured speed fell within the corresponding percentile range and estimated PSD using Welch's method (sampling rate 100 Hz; $n_{\mathrm{fft}}=128$ with 50\% overlap). PSDs were computed for each electrode and then averaged across electrodes to obtain a single spectrum per speed decile and session. To reduce the influence of the canonical $1/f$ falloff and make spectral peaks more apparent, we used a frequency-normalized spectrum by multiplying each PSD value by its corresponding frequency ($f \cdot P(f)$). Spectra were visualized from 0 to 45 Hz to avoid misinterpretation of attenuated power above the effective high-frequency content of the recordings. Finally, spectra were averaged across sessions for each speed decile, and variability across sessions is reported as mean $\pm$ SEM.

\subsubsection{Retrospective and prospective locomotion states}
\label{retrospective_and_prospective}

We extended our framework to carry out two temporal decoding tasks: forward predictions that forecast treadmill speed in the future, and backward (or reverse) predictions that infer past speeds by “flipping” the temporal order of the data, both up to a time horizon of 1000 ms. To quantify the amount of information encoded in EEG beyond what can be extracted from treadmill readings alone, we compared the performance of a neural decoder with that of the autocorrelation of the signal and a RNN trained solely on the treadmill speed signal—two baselines that exploit the highly autoregressive and smooth nature of locomotion dynamics and treadmill speeds. Dataset splits and training procedures were kept the same as in single-session experiments (Section~\ref{model_selection}). However, in this case, before initiating training, we shifted the target treadmill speed by 100, 200, 500, or 1000 ms into the past and future, effectively forcing neural decoders to estimate prior and upcoming locomotion states from the current neural activity. This analysis results in one performance estimate (correlation and R²) per session, model (EEG-based and speed-based), temporal offset and direction (forward and backward).

\subsubsection{Statistical analysis}
\label{statistical_analysis}

Decoder performance was computed per session as the Pearson correlation coefficient and the coefficient of determination R² between decoded and true treadmill speed on the test set. For each group of models (baseline decoders, transfer learning strategies, and single-region or frequency-band decoders), we first assessed normality of the session-wise scores using the Shapiro–Wilk test. Because the data was non-normal, we used non-parametric repeated-measures tests for all comparisons. Overall differences between decoding variants were assessed with a Friedman test (non-parametric repeated-measures ANOVA), applied separately to corrrelation and R². When the Friedman test was significant, we performed two-sided Wilcoxon signed-rank tests for all pairwise comparisons between models or decoding configurations (i.e., for each session we compared the performance of variant A and variant B and used the distribution of these paired differences across sessions). The resulting p-values were corrected for multiple comparisons using the Bonferroni method, and effects were considered statistically significant at a Bonferroni-corrected $p < 0.05$. The same procedure was applied consistently across all analyses involving multiple comparisons.

In the analysis of retrospective and prospective locomotion states, only two decoding variants were compared (EEG-based vs. speed-based). We therefore used a two-sided Wilcoxon signed-rank test on the paired session-wise scores for corrrelation and R² to assess differences at each temporal offset and direction. Effects were considered statistically significant at $p < 0.05$.

\section{Results}

In this study, we analyzed a large dataset of cortex-wide bilateral EEG signals sampled anterior-posterior from frontal cortex to visual cortex. These signals were synchronously recorded alongside treadmill speed from 225 sessions in 14 head-fixed rats. Speed profiles were self-controlled and varied throughout the sessions. Our central question was whether these distributed EEG signals encode enough information for continuous, online decoding of treadmill speed and, if so, which model types extract that information most effectively.

\subsection{LSTM-based recurrent neural networks provide highly accurate instantaneous decoding of treadmill speed}

\begin{figure}[htbp]
 \centering
        \includegraphics[width=1\textwidth]{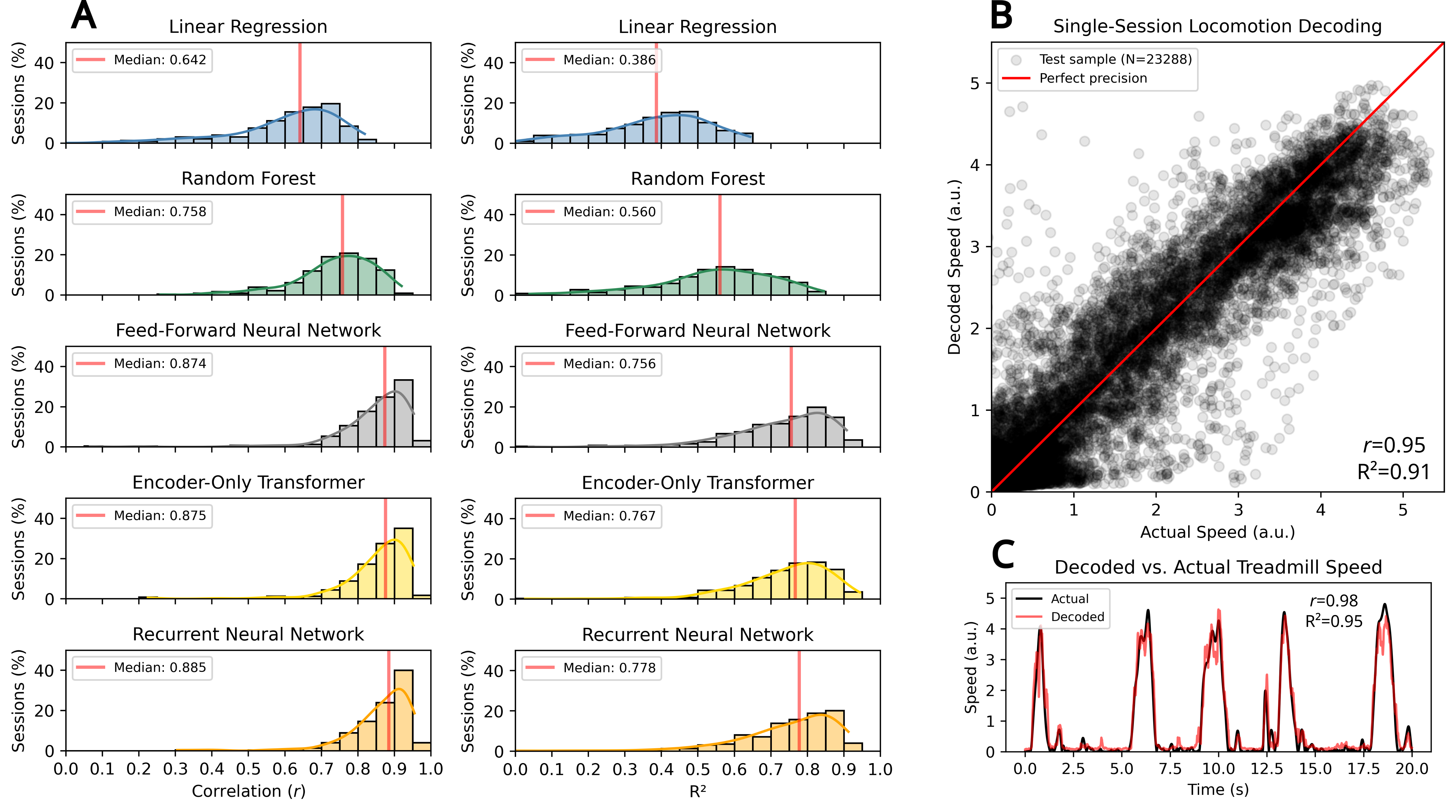}
 \caption{\textbf{Recurrent neural networks with LSTM units achieve high-precision decoding of treadmill speed on a sample-by-sample basis.} \textbf{A)} Decoding performance distribution across sessions for different machine- and deep-learning models. Each histogram represents the performance of individual sessions, measured by correlation (r, left) and coefficient of determination (R², right). The red horizontal lines indicate the median performance across sessions. \textbf{B)} Scatter plot of actual vs. decoded treadmill speeds (at 10ms intervals) on test data for one of the top 5\% best-performing sessions. Speeds are decoded using the RNN, which achieved the highest overall performance across sessions. \textbf{C)} Example speed trace from a 20-second segment of the session shown in panel B, comparing actual (in black) and decoded (in red) treadmill speeds over time.}
\label{fig:model_performance}
\end{figure}

Deep-learning methods benefit from large amounts of training data, thus, the core of our study—and the majority of our modeling effort—focused on three deep-learning architectures: a Feed-Forward Neural Network, a Recurrent Neural Network with Long Short-Term-Memory units, and an Encoder-Only Transformer with self-attention. To contextualize their performance against more traditional techniques, we also included two lightweight benchmarks that are well-known to provide reasonable baselines with little to no tuning: Linear Regression and a Random Forest Regressor. These 5 models decoded the treadmill speed from EEG activity increasingly better (\autoref{fig:model_performance}A).

The RNN significantly outperformed all other approaches, achieving a median session correlation of 0.88 and R² of 0.78 between decoded and actual speeds, across more than 13 hours of test data (see \href{https://docs.google.com/spreadsheets/d/12vc-Sul2AjtMn3D7VbMtZNOV54iBW0T1/edit?usp=sharing&ouid=104978994468990027253&rtpof=true&sd=true}{Supplementary Tables 1-3} for the results of non-parametric repeated-measures ANOVA (Friedman test, $\chi^2(4)$) and two-sided Bonferroni-corrected Wilcoxon pairwise tests; all $p < 0.0001$, except Encoder-Only Transformer vs. RNN for R² where $p = 0.0374$). In comparison, prior work using regression-based decoding of self-paced locomotion has achieved correlations up to 0.80 after post-hoc selection of the best-performing cerebellar neural ensembles \cite{muzzu2018}, and R² values of approximately 0.45 using dorsal cortex optical imaging \cite{benisty2023}. The RNN on cortex-wide bilateral EEG signals exceeded this previously reported maximal correlation of 0.80 in 82.7\% of the 225 sessions. Moreover, we observed near-perfect decoding (R² $>$ 0.90) in 5\% of the sessions, capturing almost all of the variability in treadmill speeds (see \autoref{fig:model_performance}B-C for an example session). Based on these results, we selected the RNN for the subsequent steps of this analysis. 

Next, we sought reasons why the RNN might outperform the other models. Notably, the markedly lower decoding accuracy of the simpler linear model (median correlation of 0.64, R² of 0.39) pointed to the presence of crucial relationships in the data that could not be captured by assigning a single weight per feature (Linear Regression even predicted negative speeds, see \href{https://docs.google.com/spreadsheets/d/12vc-Sul2AjtMn3D7VbMtZNOV54iBW0T1/edit?usp=sharing&ouid=104978994468990027253&rtpof=true&sd=true}{Supplementary Figure 1}). This means that models with more parameters and non-linear transformations are needed to better explain locomotion dynamics. However, the temporal structure of the data appeared less important than anticipated, as a Feed-Forward Neural Network that treated each EEG sample independently achieved only slightly lower performance to the RNN (0.87 vs. 0.88 for correlation and 0.76 vs. 0.78 for R²), which explicitly models temporal dependencies.

\subsection{Neural signature of locomotion generalizes across sessions with the same subject}

Transferring decoding performance across sessions is crucial for reducing BCI training time. If locomotion can be decoded using limited new training data—for example, by incorporating information from previous sessions—the model would require substantially less calibration during setup and could be deployed for real-world use much earlier. We therefore further evaluated the RNN model under conditions with limited or no training data from the target session. We tested six model variants grouped into three training strategies (see Methods, \autoref{fig:transfer_learning_diagram} for a diagram of the different model configurations). First, single-session models trained from scratch on either 80\% of the target session (as in the original analysis, serving as a best-case scenario baseline where large amounts of data are available) or 10\% of the data; second, zero-shot generalization, where a pre-trained model—either from the same animal (cross-session) or from a different animal (cross-subject)—is applied to the target session without any further training; and third, transfer learning via fine-tuning, where the pre-trained cross-session or cross-subject model is calibrated using only 10\% of the new session adapting its weights to the new conditions. 

\begin{figure}[htbp]
 \centering
        \includegraphics[width=0.8\textwidth]{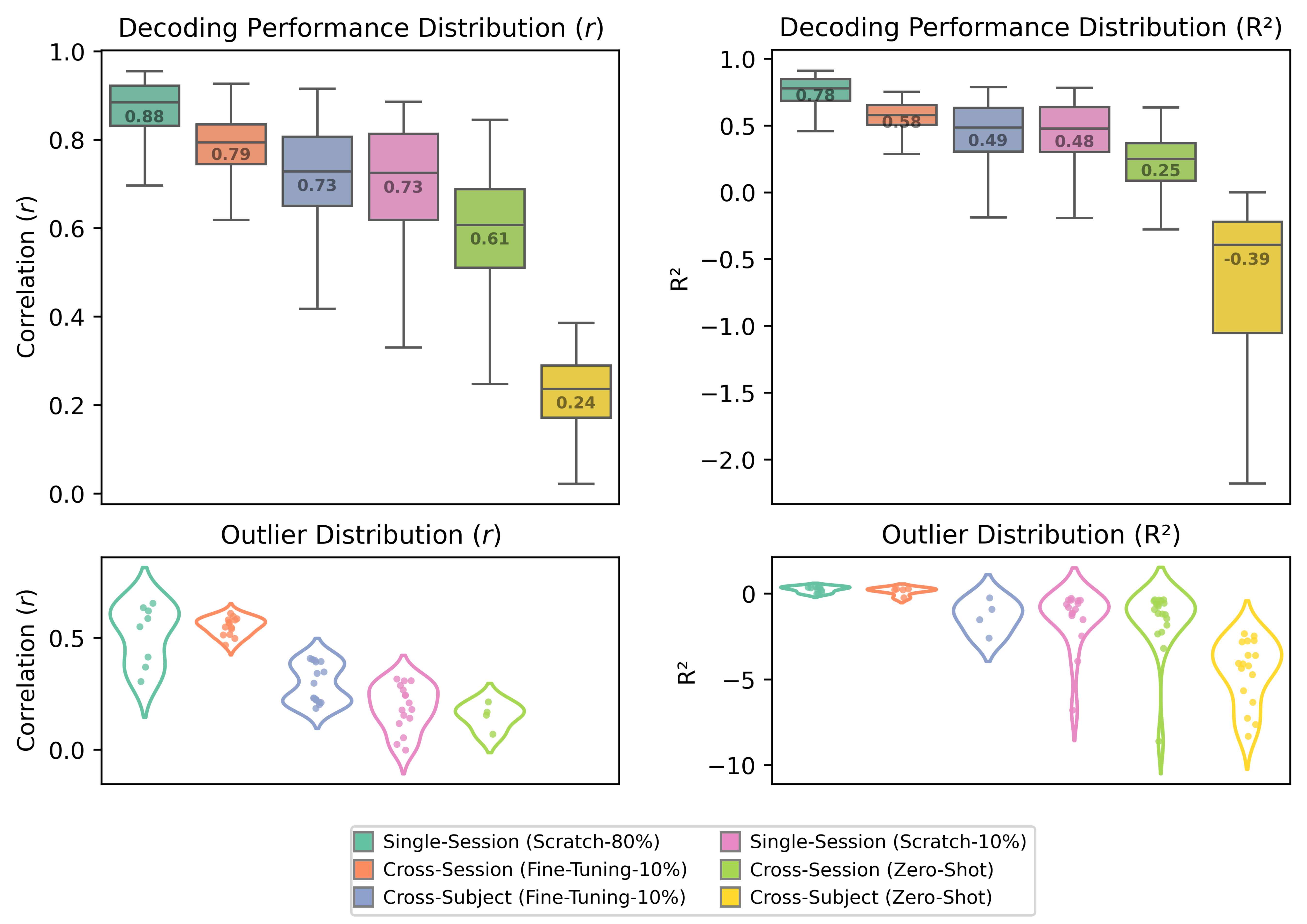}
 \caption{\textbf{Neural signatures of locomotion are conserved across sessions on multiple days within a single subject.} Decoding performance distribution for different RNN models across three training strategies: single-session from scratch (80\% training in dark green, 10\% training in pink); zero-shot—i.e., using a model trained on another session of the same animal (cross session, in light green) or from another animal (cross animal, in yellow) and fine-tuning (i.e., transfer learning) a pre-trained model (cross-session in orange, cross-subject in blue). Performance is shown for correlation (r, left) and coefficient of determination (R², right). The numbers inside the boxes show the median value of that metric. The top section displays boxplots without outliers, while the bottom section shows violin plots of the outliers separately to improve the visualization of the overall data distribution. (The cross-subject zero-shot strategy, in yellow, did not exhibit outliers in the correlation measurements.)}
\label{fig:transfer_learning_results}
\end{figure}

\autoref{fig:transfer_learning_results} summarizes the performance of every training strategy. All pairwise differences between variants were statistically significant (see \href{https://docs.google.com/spreadsheets/d/12vc-Sul2AjtMn3D7VbMtZNOV54iBW0T1/edit?usp=sharing&ouid=104978994468990027253&rtpof=true&sd=true}{Supplementary Tables 4-6} for results of the non-parametric repeated-measures ANOVA (Friedman test $\chi^2(5)$) and Bonferroni-corrected two-sided Wilcoxon signed-rank tests; all $p < 0.0001$).

Decoding precision was highest for the single-session 80\% baseline (r = 0.88, R² = 0.78). The next best results came from fine-tuning a pre-trained cross-session model (r = 0.79, R² = 0.58), followed by fine-tuning a pre-trained cross-subject model (r = 0.73, R² = 0.49) and the single-session 10\% condition (r = 0.73, R² = 0.48). Zero-shot decoding showed a significant drop in performance, though cross-session models reached a correlation of 0.61 and R² of 0.25, while cross-subject models performed worst at a correlation of 0.24 and R² of -0.39.

The partial success of cross-session zero-shot decoding (r and R² went down from 0.88 to 0.61 and from 0.78 to 0.25, respectively) and the poor performance across animals (r fell to 0.24 and R² to -0.39, respectively; indicating that the model did worse than a trivial baseline that always outputs the mean treadmill speed, by definition R² = 0) suggest that the neural signatures of locomotion are preserved to a much greater extent, across days for an individual subject, than between subjects. This may be consistent with the subject-specific nature of the signature hindering useful generalization across subjects without additional calibration. This observation is further supported by applying transfer learning via fine-tuning to both cross-session and cross-subject decoders. Fine-tuning a pre-trained model results in meaningful improvements (better performance than training from scratch on 10\% of the target session) only when this has extracted locomotion-related features from sessions with the same animal, allowing transferable information—r and R² are 0.79 and 0.58 for transfer learning compared to 0.73 and 0.48, for single-session scratch-10\%, respectively. In contrast, fine-tuning a cross-subject pre-trained model (cross-subject fine-tuning 10\%) led to similar performance as training a decoder from scratch on the available data at the start of the session (single-session 10\%; r and R² 0.73 and 0.49 vs. 0.73 and 0.48, respectively). Although the difference was statistically significant, the effect size is negligible and unlikely to be practically relevant. To the extent observed, fine-tuning appeared to help reduce the incidence and severity of outliers rather than meaningfully improve decoding performance. This suggests that, in the cross-subject setting, fine-tuning primarily acts as a regularization or stabilization step, adapting the model to session-specific statistics without conferring substantial benefits from the pre-trained representations.

\subsection{Visual cortex signals are the predominant cortical contributors to decoding locomotion speed}

\begin{figure}[htbp]
 \centering
        \includegraphics[width=1\textwidth]{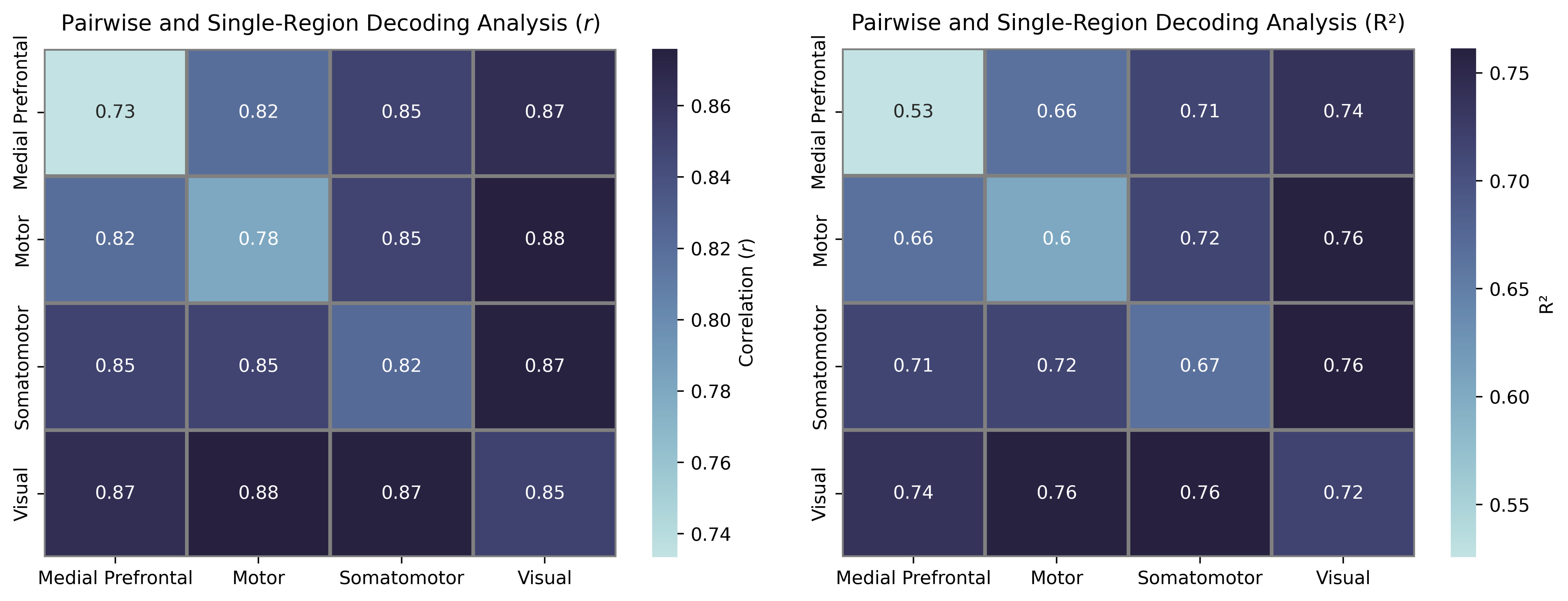}
 \caption{\textbf{Visual cortex is the predominant contributor to locomotion decoding.} Decoding performance across sessions for RNN models trained and tested on five specific brain regions (diagonal elements) and on all pairs of regions (off-diagonal elements); measured by the median correlation (r, left) and coefficient of determination (R², right).}
\label{fig:localized}
\end{figure}

Prior work on decoding locomotion has primarily targeted neurophysiological recordings in the motor, premotor, and somatosensory cortices, as well as the cerebellum \cite{rigosa2015,mirfathollahi2022,muzzu2018,fitzsimmons2009,foster2014,capogrosso2016,wang2021}. Given that the 32-channel electrode array covered the entire rat cortex, we assessed the contribution of specific cortical regions to the performance of our locomotion speed decoder. Building on the single-session analysis, we trained and tested the RNN decoder on electrodes that were placed above individual cortical regions as well as on groups of electrodes that were above different regions to determine which areas contribute the most to decoding performance. This evaluation included four individual bilateral regions (medial prefrontal, somatomotor, motor, and visual, including posterior parietal areas) and all 6 region pairs (i.e., all possible two-region combinations), each tested across all sessions and using the same training, validation, and test data splits (\autoref{fig:localized}).

Almost all pairwise differences between variants were statistically significant (see \href{https://docs.google.com/spreadsheets/d/12vc-Sul2AjtMn3D7VbMtZNOV54iBW0T1/edit?usp=sharing&ouid=104978994468990027253&rtpof=true&sd=true}{Supplementary Tables 7-9} for results of the non-parametric repeated-measures ANOVA (Friedman test $\chi^2(3)$) and Bonferroni-corrected two-sided Wilcoxon signed-rank tests). We found that using visual cortex electrodes alone performed almost as well as using all 32 electrodes (r = 0.85, R² = 0.72 for visual cortex only versus r = 0.88, R² = 0.78 for all electrodes combined, but the cortex-wide decoder is still significantly superior with $p < 0.0001$). Moreover, incorporating the visual cortex as one of a pair of cortical regions consistently boosted performance to levels nearly as high as those observed with all 32 electrodes (r $= 0.87$, R² $= 0.74$ on average). 

Decoding performance was maximized when either motor or somatomotor and visual cortex channels were combined (r $= 0.88$, R² $= 0.76$). Practically speaking, both approaches achieved the same correlation and R² as the model using all brain regions, but the cortex-wide decoder still performed significantly better ($p < 0.0001$). Decoding based on other regions, but excluding the visual cortex, resulted in clearly lower performance on average (r = 0.84, 0.84, 0.85 and R² = 0.69, 0.69, 0.72 for medial prefrontal, somatomotor and motor cortices, respectively). This suggests that the visual cortex is the primary driver of accurate decoding in this dataset, though motor areas–and especially the somatomotor system–carry substantial information alone to achieve highly precise estimations of locomotion speed. 

\subsection{Low-frequency ($<$ 8 Hz) oscillations account for most of the decoding performance}

\begin{figure}[H]
 \centering
        \includegraphics[width=1\textwidth]{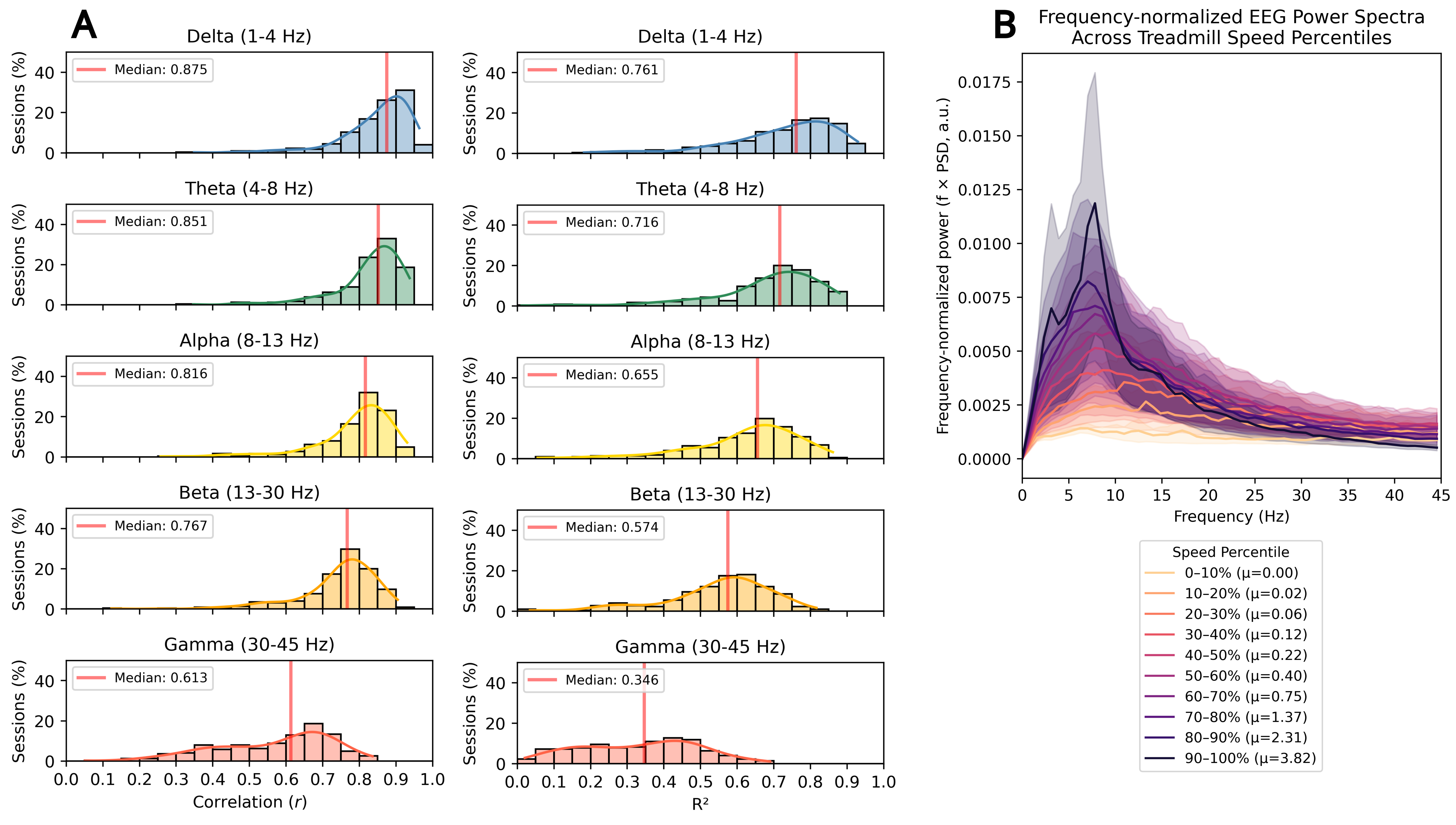}
 \caption{\textbf{Decoding performance is primarely driven by low-frequency ($<$ 8 Hz) oscillations}. \textbf{A)} Decoding performance distributions across sessions for RNN models trained and tested on isolated frequency bands. Each histogram represents the performance of individual sessions, measured by correlation (r, left) and coefficient of determination (R², right). The red horizontal lines indicate the median performance across sessions. \textbf{B)} Mean EEG spectra conditioned on speed percentile. Within each session, speed samples were binned into deciles and Welch PSDs were computed from EEG segments corresponding to each bin ($f \in [0,45]$ Hz). PSDs were averaged across channels, transformed to a frequency-normalized representation ($f \cdot P(f)$) to flatten the $1/f$ slope, and then averaged across sessions (shaded areas: SEM).}
\label{fig:frequencies}
\end{figure}

The broadband EEG field potential encompasses many oscillatory components that have distinct relationships to different transmembrane currents, neuronal microcircuit motifs, and cross-region neuronal networks \cite{buzsaki2012,buzsaki2004}. Given that different oscillatory activity in distinct frequency bands arises from transmembrane currents associated with specific ion channels and neurotransmitters, some bands may be more informative than others for decoding actions associated with particular neurophysiological processes \cite{buzsaki2012, polack2013}. Therefore, we evaluated the decoder on isolated frequency bands by training and testing the RNN separately on each of five bands (delta, 1–4 Hz, theta, 4–8 Hz, alpha,  8–12 Hz, beta, 12–30 Hz, and gamma 30 Hz highpass, see \autoref{fig:frequencies}A).

All pairwise differences between variants were statistically significant (see \href{https://docs.google.com/spreadsheets/d/12vc-Sul2AjtMn3D7VbMtZNOV54iBW0T1/edit?usp=sharing&ouid=104978994468990027253&rtpof=true&sd=true}{Supplementary Tables 10-12} for results of the non-parametric repeated-measures ANOVA (Friedman test $\chi^2(4)$) and Bonferroni-corrected two-sided Wilcoxon signed-rank tests). We found that delta and theta bands produced similar decoding results, but not significantly better, to those of a model with access to the entire spectrum (r = 0.88, R² = 0.76 and r = 0.85, R² = 0.72, with $p < 0.0001$ for delta vs. full-band and theta vs. full-band, respectively). We observed this pattern not only in the decoding results but also in the grand-averaged, frequency-normalized power spectra conditioned on treadmill speed percentiles, pooled across all animals and sessions (\autoref{fig:frequencies}B). Specifically, power increases with speed, with the largest differences at low frequencies. At the highest running speeds, a distinct peak emerges around 8 Hz, near the upper edge of the theta band. It is plausible that a sufficiently complex algorithm, such an RNN, could model this relationship for decoding purposes, even in end-to-end approaches using all frequencies.

\subsection{Predictive decoding of treadmill speeds up to 1 second in the future}

\begin{figure}[H]
 \centering
        \includegraphics[width=0.8\textwidth]{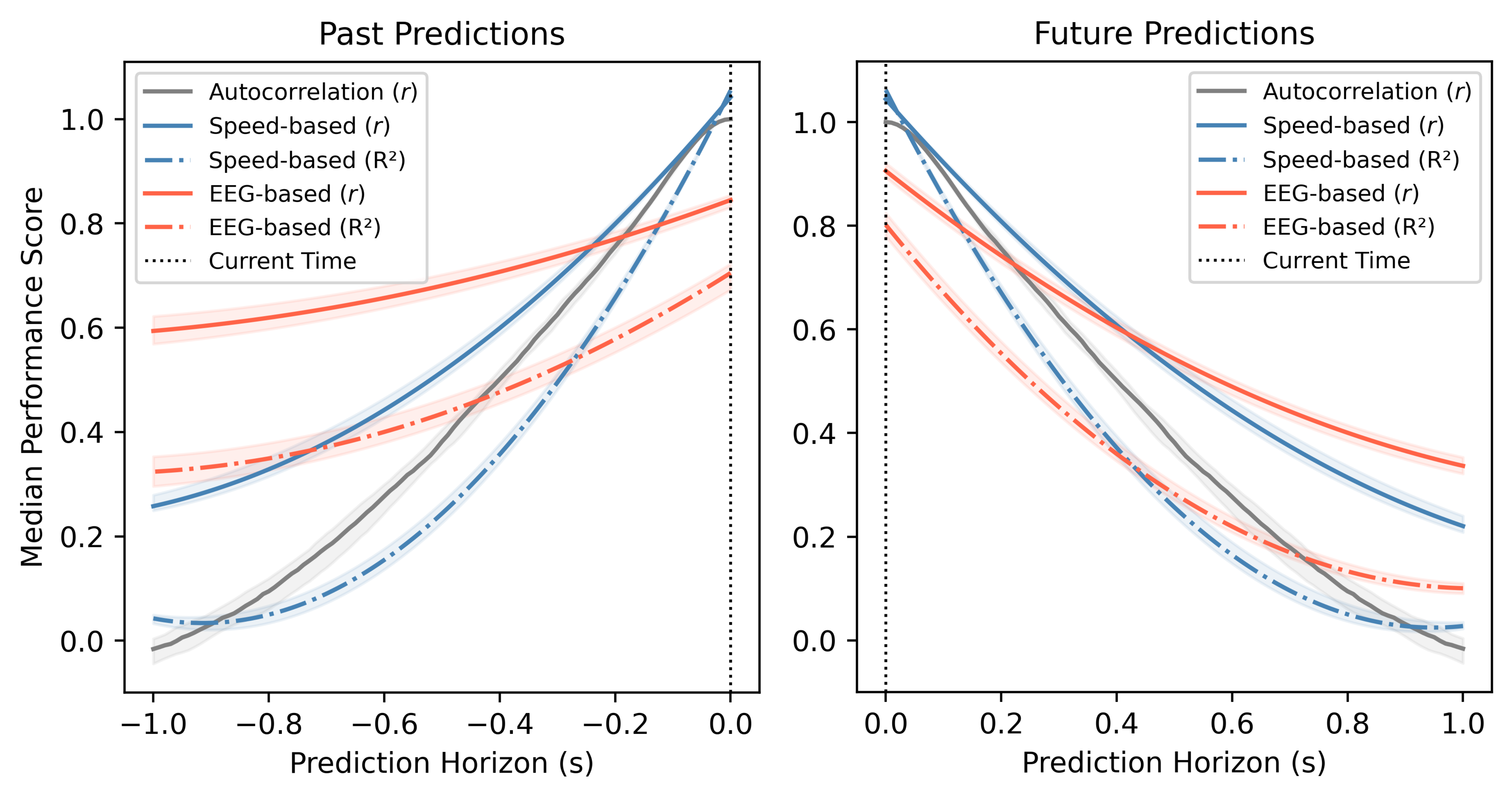}
 \caption{\textbf{EEG encodes information about previous and future treadmill speeds beyond what can be estimated from treadmill readouts alone}. Decoding performance across sessions for EEG-based (red), speed-based (blue) and autocorrelation (gray) models estimating previous speeds (left) and future speeds (right). Models were evaluated at discrete time horizons (0, $\pm$100, $\pm$200, $\pm$500, and $\pm$1000 ms), with 95\% confidence intervals computed from bootstrapped medians across sessions. To aid visualization and analysis, a second-order polynomial was fit to these points, resulting in the smooth curves shown (see \href{https://docs.google.com/spreadsheets/d/12vc-Sul2AjtMn3D7VbMtZNOV54iBW0T1/edit?usp=sharing&ouid=104978994468990027253&rtpof=true&sd=true}{Supplementary Tables 13-14} for the actual performance on discrete time horizons). The signal autocorrelation was calculated for all lags, in 10 ms increments, up to $\pm$1000 ms without polynomial approximation.}
 \label{fig:predictions}
\end{figure}

A real-world BCI for mobility requires, not only instantaneous real-time neural decoding of walking speed, but also the ability to predict future speeds to efficiently plan exoskeleton movements. Therefore, we tested the extent to which it is possible to accurately decode future speeds from present EEG activity. We also compared the accuracy of this prediction to the same time lags in the past, under the assumption that decoding of planning and intention formation would be less feasible than decoding a deterministic past. We quantified how much temporal locomotion-related information is encoded in neural data by benchmarking the EEG-based RNN decoder's performance against two baselines: (1) the autocorrelation of the treadmill speed signal across different lags and (2) an RNN with the same architecture as the decoder but trained solely on treadmill speeds with no access to brain activity (speed-based). Both analyses were run forward and backward at lags up to 1000 ms. We summarized this analysis by comparing the median performance of the models across all sessions as a function of the decoding horizon, highlighting differences between EEG-based and speed-based models (\autoref{fig:predictions}).

Across all delays and for both forward and backward directions, the EEG-based and speed-based decoders differed significantly in performance (all two-sided Wilcoxon signed-rank tests, $p < 0.0001$; see \href{https://docs.google.com/spreadsheets/d/12vc-Sul2AjtMn3D7VbMtZNOV54iBW0T1/edit?usp=sharing&ouid=104978994468990027253&rtpof=true&sd=true}{Supplementary Table 15}). Our results indicate that treadmill speed readouts alone allow for fairly reliable estimates of past and future speeds over short time scales (see the Autocorrelation (r) curve in black in both panels of \autoref{fig:predictions}). In fact, the autocorrelation is higher than the EEG-based correlation until  $\sim 200$ ms into the past and future. But the autocorrelation decays quickly and symmetrically in the past and future directions. Similarly, the speed-based correlation and R² (in solid and dash-dotted blue lines, respectively, in \autoref{fig:predictions}) are higher than EEG-based until $\sim 300$ ms into the past and $\sim  400$ ms into the future; though again such speed-based predictions decay much faster than the EEG-based ones. In contrast, EEG-based predictions begin lower than the autocorrelation and speed-based predictions but decay at a much lower rate. They also exhibit an asymmetry: predictions of past speeds decay much more slowly, reaching a correlation of 0.59 and R² of 0.32 at -1000 ms; whereas future speed predictions decay faster, with a correlation of 0.34 and R² of 0.10 at +1000 ms into the future.

These results suggest that locomotion-related information persists in EEG, enabling reasonably accurate estimates of recent speeds. In contrast, predictive information appears less accessible, indicating that prospective representations are not as well reflected.

\section{Discussion}

We set out to test the extent to which self-paced locomotion speed can be decoded from cortex-wide EEG in rats on a non-motorized treadmill. By treating locomotion as a continuous behavior—without reliance on trial markers or epochs—we trained an LSTM-based RNN on session-long, behaviourally rich datasets, achieving a median correlation of 0.88 and explaining 78\% of the variance (R²) in treadmill speed across 225 sessions from 14 animals. To our knowledge, this represents a level of precision not previously reported for self-paced locomotion speed decoding and surpasses earlier regression-based work, which has reported correlations up to an optimistic upper bound of 0.80 using invasive single-neuron recordings and post hoc selection of the best-performing neural ensembles \cite{muzzu2018, benisty2023}.

Three factors appear to have contributed to this improvement in decoding accuracy. First, the breadth of our dataset—both in duration (the 225 sessions spanned 133 hours of recording) and in cortical coverage (32 electrodes covering the visual, motor, somatomotor, and medial prefrontal cortices)—provides the RNN with a diverse training distribution that captures the natural variability of rodent locomotion. Second, self-paced locomotion introduces richer dynamics than induced, fixed-speed paradigms, potentially exposing the model to non-linear relationships that simpler tasks might not reveal. Third, deep architectures are inherently better suited than linear models for mapping high-dimensional feature spaces onto time-varying states \cite{roy2019}. In our dataset, moving from Linear Regression to an RNN recovered $\sim64\%$ of the variance left unexplained by the linear approach, underscoring the complexity and nonlinearity of the underlying neural-locomotion mapping. Even non-recurrent sequential models—such as a Feed-Forward Neural Network or an Encoder-Only Transformer, the latter now dominant in sequence modeling problems—achieved decoding performance that was slightly worse than that of the RNN.

Our analysis also highlights a major contribution from  posterior cortical electrodes, including those covering the visual and parietal cortex \cite{paxinos2013}, that is well documented physiologically. Several studies have shown that locomotion modulates activity in the rodent primary visual cortex, enhancing visually evoked responses and altering excitability even without visual input \cite{saleem2013,liska2024,erisken2014,flossmann2020,vinck2014}. Furthermore, this modulation includes more than a twofold increase in visually evoked firing rates when an animal transitions from rest to running, indicating a shift in brain state \cite{niell2010}. In our dataset, the rat locomotion was a combination of spontaneous running during inter-trial intervals, as well as choice-related running past a distance threshold in a Go/NoGo visual discrimination task. In both cases, running speed, which was the decoding target, was freely chosen by the rat in a volitional choice paradigm. Nevertheless, we considered whether the inclusion of data from a visual Go/NoGo task might explain why information in the visual cortex was, of all cortical signals, the most relevant for decoding. But there are two reasons to think that the information in the visual cortex goes beyond just a reflection of the visual cue. First, the visual cue just informed the animals whether they need to move a certain distance along the treadmill to get a reward or stay still. But the self-paced nature of the task meant that the speed profile remained up to the animals and was indeed variable throughout the sessions (\href{https://docs.google.com/spreadsheets/d/12vc-Sul2AjtMn3D7VbMtZNOV54iBW0T1/edit?usp=sharing&ouid=104978994468990027253&rtpof=true&sd=true}{Supplementary Figure 1}). Second, we estimated the average reaction time—from stimulus onset to movement initiation—to be 387 ms, which falls within the range reported in the literature for similar experimental paradigms, where rats have been shown to integrate brief visual cues over 300–800 ms before acting \cite{shevinsky2019, reinagel2012}. Such latencies make it unlikely that transient, stimulus-locked responses alone could account for the smooth, minimal-lag trajectories lasting several seconds that our model reproduces. Indeed, restricting the input to only the electrodes over motor or somatomotor areas still achieved 0.78 to 0.82 correlation and R² of 0.60 to 0.67 (compared to r = 0.85, R² = 0.72 for visual cortex; hence still capturing up to 93\% of the variance explained by the visual cortex), confirming that motor and adjacent cortices alone carry substantial locomotion information.

We found that delta and theta band features were dominant in our decoder—and using those two bands in isolation nearly matched the performance of the model when trained on the entire  spectrum.  The importance of these low frequency field potential oscillations are not surprising given the large body of evidence linking them to gait and arousal. The literature highlights prominent delta-band modulation during periods of low activity, such as when animals are stationary or moving slowly \cite{schultheiss2019}. Such conditions were frequently observed in our sessions between bouts of running. In addition, studies of rodent hippocampal field potentials have shown that both the frequency and power of the theta rhythm increase with positive acceleration, a relationship that has been used to estimate locomotion speed \cite{kropff2021,shin2001,fernandez2024,sheremet2018}. Such high theta oscillations were also later observed in human medial temporal lobe field potentials during real-world (but not virtual) locomotion \cite{aghajan2016}. Although these studies support the long-standing link between locomotion and theta band field potential oscillations in the hippocampus, work with head-fixed mice has reported that locomotion unmasks a narrow 8-9 Hz theta peak in the primary visual cortex local field potential that is virtually absent during immobility, mirroring the frequency content and cortical location that we observed in our dataset \cite{niell2010,ayaz2013}. Collectively, this evidence suggests that our model may be exploiting these relationships for decoding.

Despite identifying a clear neural correlate to decode locomotion, we found that this pattern did not translate into a universal neural signature of locomotion across animals. Transfer learning analyses showed that, while a pre-trained decoder generalized well across sessions within the same animal, performance dropped substantially when applied to sessions with different animals. This suggests that subject-specific calibration is essential. Note, however, that fine-tuning on just 10\% of the data raised the performance dramatically, indicating that decoders trained on other subjects, and certainly those trained on other sessions of the same subject, can be quickly fine-tuned to boost performance. The ability to rapidly and accurately calibrate decoders to the latest neural state is particularly valuable for real-time BCI experiments, where models are often briefly calibrated at the start of a session, with only a small amount of subject-specific data. Thus, reusing models trained on previous sessions or other subjects, combined with minimal fine-tuning, can reduce the amount of calibration required and streamline deployment in live, closed-loop systems. The opportunities that this technique offers are widely acknowledged in the field and are a key reason transfer learning and fine-tuning have become a vibrant research frontier in the BCI community, driving progress across diverse applications—including locomotion decoding—and attracting growing interest from deep-learning practitioners in recent years \cite{quiles2023,peterson2020,fu2025,tortora2020}.

A novel outcome of our study is the demonstration that cortical signals encode forthcoming and previous locomotion states beyond what can be predicted from treadmill readings alone. While signal autocorrelation and speed-based autoregressive models produce highly accurate short-term predictions of both past and future speeds, their performance degrades quickly and symmetrically in both directions over longer time horizons. In contrast, EEG-based estimates outperform speed-based models for horizons exceeding $\pm$300--400 ms and exhibit a clear asymmetry, with past speeds inferred more accurately than future speeds. This pattern suggests that locomotion-related information leaves residual traces in ongoing cortical dynamics, enabling reconstruction of recent speeds, whereas prospective speed information—which depends on planning- or intention-related processes and is partly constrained by the nature of the Go/NoGo discrimination task—is less strongly reflected in the EEG. Overall, these findings point to the idea that neural signals provide a richer temporal embedding of locomotion, offering potential advantages for real-time motor control applications that extend past direct kinematic measurements.

Although our online framework attains state-of-the-art precision, several practical considerations must be addressed before deployment in real-time settings. First, the decoder outputs low, non-zero speeds during periods of immobility (\href{https://docs.google.com/spreadsheets/d/12vc-Sul2AjtMn3D7VbMtZNOV54iBW0T1/edit?usp=sharing&ouid=104978994468990027253&rtpof=true&sd=true}{Supplementary Figure 1}). Consequently, integrating such spurious values would translate into unintended treadmill movement. Implementing a speed threshold that clips values below a behaviourally defined rest level to zero would mitigate this risk. Second, the raw decoder output is less smooth than natural gait. Applying causal gaussian or exponential filters would improve smoothness but introduce latency. Compensating for that delay with parallel predictions of future speeds—derived either from an additional neural decoder, an autoregressive speed model, or their fusion in a multimodal approach—could preserve responsiveness. Evaluating these engineering refinements, alongside safety interlocks, constitutes an important avenue for future work.

\section{Conclusions}

Our study demonstrates that non-invasive, cortex-wide EEG in rodents can reliably support continuous decoding of self-paced locomotion, broadening a literature that has predominantly emphasized invasive recordings, discrete gait states, or externally controlled speeds. Using large-scale recordings that capture natural variability in locomotion and non-linear deep-learning models, we show that distributed cortical activity—including prominent contributions from visual areas and low-frequency oscillatory structure—supports accurate reconstruction of voluntary speed dynamics, requiring in many cases only modest calibration of pre-trained models to individual subjects via transfer learning. This places our EEG-based decoding framework between precision-focused intracranial rodent studies and broader translational non-invasive BCI approaches, highlighting the utility of cortex-wide signals for asynchronous decoding in naturalistic, continuous contexts.

\ack{We thank the following members of Nelson Totah’s laboratory for contributing to the dataset: Ryo Iwai, Dmitrii Vasilev, Isabel Raposo, Negar Safaei, Lilian de Sardenberg Schmid.}

\funding{This work was supported by the Helsinki Institute of Life Science (HiLIFE), University of Helsinki (to NT), and by the Research Council of Finland (Decision \#335930 to NT). This publication was also made possible through the support of a joint grant from the John Templeton Foundation and the Fetzer Institute (to UM). The opinions expressed in this publication are those of the authors and do not necessarily reflect the views of the funders.}

\coi{The authors declare no competing interests.}

\roles{
Conceptualization: AM, UM, NT.
Methodology: AM, UM.
Software: AM.
Validation: AM, UM, NT.
Formal analysis: AM.
Investigation: NT.
Resources: NT, UM.
Data curation: AM, NT.
Writing -- original draft: AM.
Writing -- review and editing: AM, NT, UM.
Visualization: AM.
Supervision: UM, NT.
Project administration: UM, NT.
Funding acquisition: UM, NT.}

\data{The datasets analyzed in this study are openly available at the following \href{https://doi.org/10.23729/fd-375d1fcc-a405-3519-8d8c-47d96cfebb65}{DOI}.}

\suppmat{Supplementary figures and tables for this manuscript are available online at 
\href{https://docs.google.com/spreadsheets/d/12vc-Sul2AjtMn3D7VbMtZNOV54iBW0T1/edit?usp=sharing&ouid=104978994468990027253&rtpof=true&sd=true}{this link}.}

\newpage

\printbibliography

\end{document}